\documentclass[sigconf]{acmart}

\usepackage{booktabs} % For formal tables
\usepackage{multirow}

\setcopyright{rightsretained}
\acmConference[ACMMM'18]{ACM Multimedia conference}{October 2018}{Seoul, South Korea}
\acmYear{2018}
\copyrightyear{2018}

\acmArticle{4}
\acmPrice{15.00}

\begin{document}
\title[FusedLSTM and Kernel similarity learning for CBVRP]{FusedLSTM at ACMMM-2018 CBVRP Challenge:\\ Fusing frame-level and video-level features for Content-based Video Relevance Prediction}
% \titlenote{Produces the permission block, and
%   copyright information}
% \subtitle{Challenge submission report}
% \subtitlenote{The full version of the author's guide is available as
%   \texttt{acmart.pdf} document}

\author{Yash Bhalgat}
% % \authornote{Dr.~Trovato insisted his name be first.}
% \orcid{0001-7775-6250}
\affiliation{\institution{University of Michigan, Ann Arbor}}
\email{yashsb@umich.edu}

% The default list of authors is too long for headers.
% \renewcommand{\shortauthors}{B. Trovato et al.}

\begin{abstract}
This paper describes two of my best performing approaches on the Content Based Video Relevance Prediction challenge. In the FusedLSTM based approach, the inception-pool3 \citep{inception} and the C3D-pool5 \citep{c3d} features are combined using an LSTM and a dense layer to form embeddings with the objective to minimize the triplet loss function. In the second approach, an Online Kernel Similarity Learning \citep{OMKS} method is proposed to learn a non-linear similarity measure to adhere the relevance training data. The last section gives a complete comparison of all the approaches implemented during this challenge, including the one presented in the baseline paper \citep{cbvrp-acmmm-2018}.

\end{abstract}


%
% The code below should be generated by the tool at
% http://dl.acm.org/ccs.cfm
% Please copy and paste the code instead of the example below.
%



\maketitle

\section{Introduction}
Personalized recommendations have been the core focus of a major proportion of the algorithms in Information Retrieval. Video recommendation systems have gained increasing importance in both the academia and industry, in the light of the current explosive growth of popular services like YouTube, Hulu, Netflix, Twitch, etc. in general, these systems use collaborative filtering methods with intrinsic assumptions about the availability of the users' past watching behaviors or relevance feedback (ratings, reviews) on the videos \citep{collab1,collab2}. A lot of video recommendation systems are based on just the meta-data, titles or the tags in a video \citep{multimodal_feedback}. In addition to these features, YouTube recommendations try to estimated the expected time the user spends watching a video \citep{time_watched_youtube}.

The aim of this challenge was to be build a system which will be able to provide recommendations based solely on the implicit visual content in the videos. The motive of this constraint is to tackle the "cold-start" problems in recommendation systems, which occurs due to the lack of the behavioral data on the users on a video which is newly added to the database. 

\section{Data} \label{data}

In this challenge, to protect the privacy of the users in the collected data, we were provided with pre-extracted features of the videos \citep{cbvrp-acmmm-2018}. Two kinds of features were included in the dataset - 
\begin{itemize}
\item \textbf{Inception-pool3}: Each frame is passed through the inception network \citep{inception} and the output of the Pool-3 layer is used. 
\item \textbf{C3d}: The video is passed through a trained 3D convolutional neural network \citep{c3d}. The obtained embeddings are expected to contain sufficient information for video retrieval.
\end{itemize}

There are two tracks of videos, namely TV-shows and movies-trailers. The distribution of the datasets was as follows:
\begin{itemize}
\item \textit{Movies}: training set (4,500 movies), validation set (over 1000 movies), and testing set (4,500 movies)
\item \textit{Shows}: training set (3,000 shows), validation set (over 800 shows), and testing set (3,000 shows)
\end{itemize}

\section{Preliminaries}
For the task of relevance prediction, it's useful to use the idea of a relevance function $r$ \citep{deep_metric_learning}. In this task, for a set of samples $\mathcal{P}$, if we have three videos $p, p^+, p^-$, we want to be able to say that $r(p, p^+)>r(p, p^-)$. $p$ is often referred as the \textit{anchor}.

\subsection{Triplet loss function}
Given the nature of the training data, the objective of the loss function is to learn representations such that the "similarity" between the anchor and the positive video is maximized and minimizes the similarity measure between the anchor and the negative. This constraint can be formulated as:
$$ S(p,p^+) > S(p,p^-) + margin $$
where the margin hyperparametes is tuned based on the similarity function used or the application.

The triplet loss can be written as:
$$ \mathcal{L} = max(0, margin + S(p,p^-) - S(p,p^+)) $$
This formulation encompasses the above constraint. When the constraint is satisfied, the loss becomes zero and the corresponding triplet does not contribute to the training henceforth.

\subsection{Similarity measures} \label{similaritykernels}
For generality of notation, let $\Delta(p,q)$ denote the \textit{l-2} distance between two embeddings $p$ and $q$. In this challenge, the following similarity kernels were utilized in the experiments:

\begin{enumerate}
\item \textbf{RBF kernel}: $S(p,q) = exp(-\frac{\Delta(p,q)}{\gamma \sigma^2})$
\item \textbf{Shifted cosine}: $S(x,y) = 0.5+0.5\langle x,y \rangle$
\item \textbf{Softmax}: \citep{deep_metric_learning} For a triplet $(p, p^+, p^-)$, we define the similarities as 
$$ S(p, p^+) = \frac{e^{-\Delta(p,p^+)}}{e^{-\Delta(p,p^+)}+e^{-\Delta(p,p^-)}}, S(p, p^-) = \frac{e^{-\Delta(p,p^-)}}{e^{-\Delta(p,p^+)}+e^{-\Delta(p,p^-)}}$$

\end{enumerate}

\subsection{Regularization:}
With the rbf and softmax kernels, quick over-fitting is observed, because the network tends to form very large embeddings such that $e^{-\Delta(p,p-)} \to 0$. To avoid the explosion of the norms of the embeddings, a regularization is applied:

$$ \mathcal{L} = max(0, margin + S(p,p^-) - S(p,p^+)) + \lambda\sum_{p, p^+, p^-} \lVert h(p) \rVert$$

\subsection{Triplet selection and mirroring}
In the training data, for each anchor video, we are provided a \textit{ranked} list of videos relevant to the anchor. So, I assume that all the other videos not in the list are \textit{irrelevant} to the anchor video. (Note that a triplet is denoted as a tuple $(p, p^+, p^-)$ where $p$ and $p^+$ are related and $p$ and $p^-$ are not related.) A straightforward approach would be to choose the anchor video as $p$, all the videos in the relevance list as $p^+$ and all the videos not in the list as $p^-$. But this leads to over-fitting since the same anchor video appears repeatedly in a lot of triplets.

\subsection*{Anchor Point Mirroring}
To avoid this problem, we note that if $(p, p^+, p^-)$ is a valid triplet, $(p^+, p, p^-)$ is also a valid triplet. Hence, we can randomly choose any two videos from the relevance list (including the anchor video) as $p$ and $p^+$. And choose $p^-$ as before. This gives a huge improvement in the variability of data, hence reducing the risk of overfitting.

\section{Proposed Approaches}
\subsection{Kernel based Similarity Learning}
It has been proven that non-linear functions are capable of learning complex patterns. Hence, this method builds upon the online learning framework OMKS \citep{OMKS} which tries to learn a non-linear similarity measure between two video features. In this method, for a given kernel function $\kappa(\cdot, \cdot)$, we try to learn a linear operator $L$, s.t. the similarity is defined as: $S_L(p,q) = \langle \kappa(p, \cdot), L[\kappa(q, \cdot)] \rangle$

As proposed in OMKS, at each iteration $i$, the learning is performed as follows:

\begin{equation} \label{eq:kernel1}
\tau_i = min\left\{C, \frac{max(0, margin + S_{L_{i-1}}(p_i,p_i^-) - S_{L_{i-1}}(p_i,p_i^+))}{\kappa(p_i,p_i)[\kappa(p_i^+,p_i^+)-2\kappa(p_i^+,p_i^-)+\kappa(p_i^-,p_i^-)]} \right\}
\end{equation}

where, $\tau$ is the learning coefficient. 

The similarity between two embeddings at the point $i$ in time is calculated as:

\begin{equation} \label{eq:kernel2}
S_{L_{i}}(p,q) = \kappa(p,q)+\sum_{k=1}^{i} \tau_k \kappa(q,p_k)(\kappa(p,p_k^+)-\kappa(p,p_k^-))
\end{equation}

\subsection*{Feature Formation}
\textit{Note: This section also carries on the to the next FusedLSTM approach mentioned in section \ref{fusedLSTM}.}

As mentioned in section \ref{data}, we have two kinds of features. The frame-level features of size $n\_frames \times 2048$ formed by passing each frame of the video through a Inception-V3 network and a 512 length video-level vector formed as an output of passing the video through a 3d convolutional network.

In one of the experiments, for the frame-level features, a mean is calculated for each frame and used as the final vector. To encode more temporal information about the video, later on 6 other statistical measures were recorded for each feature along the time-dimension, namely \textit{max}, \textit{min}, \textit{median}, \textit{25\% quartile}, \textit{75\% quartile} and \textit{standard deviation}. Some pooling was also used to reduce the feature size for faster learning. The feature obtained with this was then concatenated with the C3D feature vector to give the final representation.

Furthermore, for another set of experiments, delta and delta-delta features are also incorporated. This does not give much improvement, hence these are not discussed in this paper.

\subsection*{Implementation}

Equations \ref{eq:kernel1} and \ref{eq:kernel2} forms the core of kernel similarity learning. Different kernel functions can be used to improve upon the simple cosine similarity as described in \ref{similaritykernels}.

In my online learning, the number of triplets (training samples) are of the order of $10^7$. So, the \textit{testing} step using \ref{eq:kernel2} involves multiplying 3 matrices of the same order. Hence, the matrix multiplication was implemented in CUDA to reduce the big matrices into smaller blocks which can be handled by the GPU cores. The computation time was reduced from 16 hours on CPU to 766 seconds on GPU for \#triplets = $10^7$.

\begin{figure*}
\includegraphics[width=\textwidth]{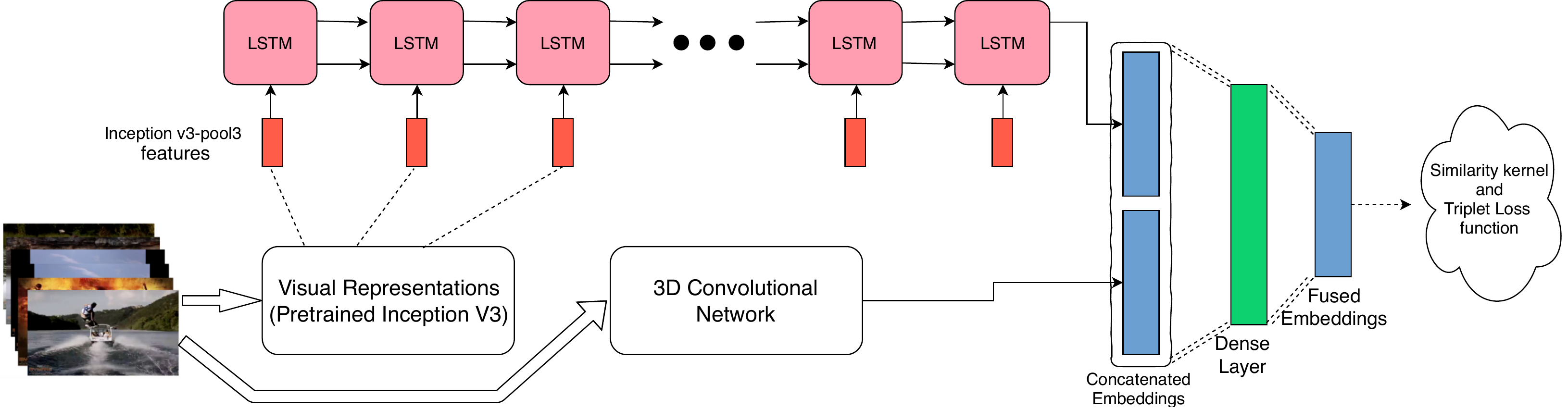}
\caption{Close-up of a gull\label{fig:fusedLSTM}}
\end{figure*}

\subsection{FusedLSTM Representation Learning} \label{fusedLSTM}
The inference step in the kernel learning method is computationally intensive. Hence, instead of trying to learn the similarity metric $S(p,q)$, if we learn an intermediate representation mapping $h(\cdot)$, we can \textbf{pre-compute} the embeddings and the inference step will be an $O(1)$ calculation (keeping the embedding size constant). A FusedLSTM approach is proposed where the \textit{inception} (sequential) and \textit{c3d} features are combined using a Dense layer to form embeddings.

\subsection*{Architecture}
The scheme proposed in Figure \ref{fig:fusedLSTM} consists of a variable-length single-layer LSTM, where the output of each cell passed on to the next cell. The frame-level features obtained from the Inception-V3 network are passed as inputs to each of these cells. The LSTM captures the temporal relationships in the frames which contribute to the relevance prediction. To utilize the video-level information, the output of the last cell of the LSTM is concatenated with the video-level (C3D) feature. This concatenated embedding is passed through a fully connected layer to give us the final \textit{\textbf{fused embeddings}}.

\begin{figure}[H]
\includegraphics[width=0.45\textwidth]{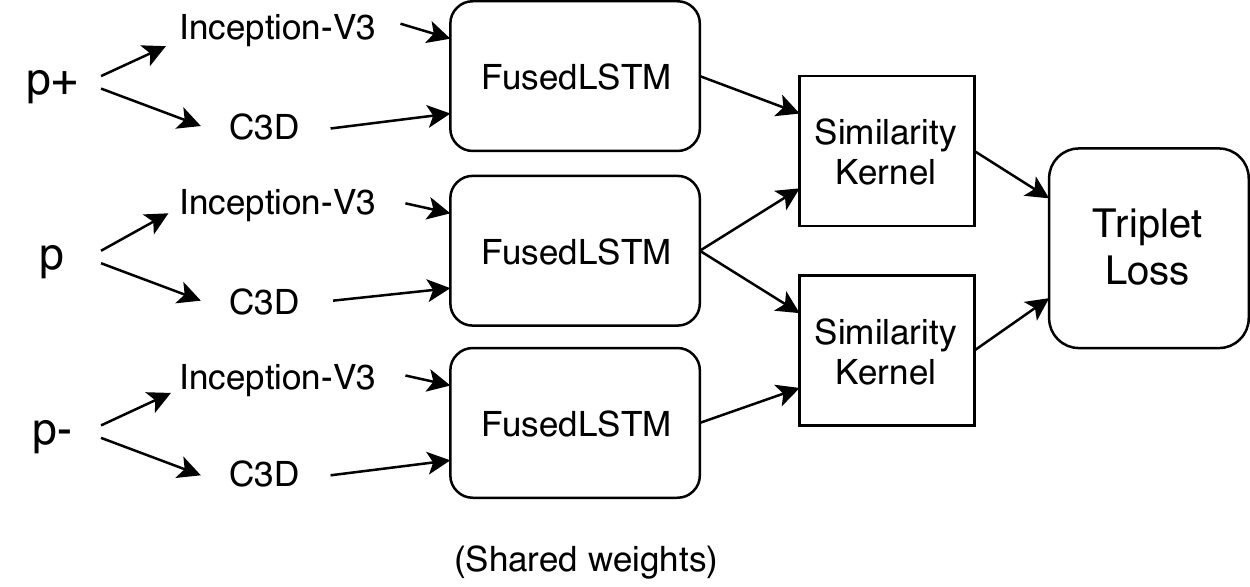}
\caption{Triplet Net architecture\label{fig:TripletNet}}
\end{figure}

As shown in Figure \ref{fig:TripletNet}, three instances of this FusedLSTM network are combined to form a higher level architecture called the \textit{Triplet Network} \citep{deep_metric_learning}. For every triplet, the video-level and frame-level features for each video are passed through the modules respectively to give the embeddings as the last layer of the FusedLSTM. The similarities $S(p,p^+)$ and $S(p,p^-)$ are calculated based on these \textit{fused-embeddings} obtained for the videos. The triplet-loss function is applied to these outputs as described earlier. 

\subsection*{Implementation and Training}
This approach was developed in PyTorch. \footnote{The source code can be provided if required.} As described earlier, the loss function is such that $S(p,p^+)$ is maximized and $S(p,p^-)$ is minimized. The loss function can also be seen as trying to reach a state where $S(p,p^+) \to 1$ and $S(p,p^-) \to 0$ (-1 in case of cosine similarity). Training was done using a simple Adam's optimizer and the model was trained for \textasciitilde $15 \times 10^6$ iterations, where each triplet sample is seen \textit{only once}.

\section{Experiments and Results}
In this section, a complete analysis and comparison of all the attempted methods and hyperparameters is provided.

\subsection{Experiments}

To reproduce the results in the baseline paper \citep{cbvrp-acmmm-2018} provided by the authors of the CBVRP challenge, a triplet network with a single fully-connected layer was developed for relevance learning.

To improve upon the validation results, several approaches were developed to tackle this challenge, which can be summarized as: 
\begin{enumerate}
\item Bilinear similarity metric learning (OASIS algorithm) \citep{OASIS}
\item Kernel similarity learning (based on OMKS) \citep{OMKS}
\item 2-layer Neural Network
\item FusedLSTM based Triplet Network
\end{enumerate}

OASIS algorithm for learning bilinear similarity \citep{OASIS} was implemented, but it performed poorly compared to the baseline. To improve the performance, the non-linear version of this algorithm OMKS (Online Metric Kernel Learning) was implemented. This method performed better than the baseline on the validation set as can be seen in the Tables 1 and 2.

As the next set of experiments, the \textit{Triplet Network} with different architectures for the embedding-net was used. The tunable parameter in these experiments was the number of epochs and embedding size. Initially, a simple 2-layer neural network with varying embedding sizes of 128, 256 and 512 was used. Choosing an embedding size more than 256 led to overfitting on the data. This model gave \textit{at par}, but not better results than the challenge baseline. So, to improve upon this the FusedLSTM approach was used. An embedding size of 256 gave the best results on the validation dataset on both movies and shows as can be seen in the tables below. 

In all these experiments, different kernel functions were tried - RBF, cosine and softmax as defined in section \ref{similaritykernels}. Metrics used for evaluation in this challenge are \textit{hit-rate} and \textit{recall} based on top K predictions as described in the baseline paper \citep{cbvrp-acmmm-2018}. The challenge is evaluated on the basis of \textit{hit-rate@30} and \textit{recall@100}. The FusedLSTM approach performs better than the baseline (hit-rate=\textbf{0.510} and recall=\textbf{0.175}) with all the kernels with the softmax kernel achieving a \textit{hit-rate@30} of \textbf{0.483} and \textit{recall@100} of \textbf{0.205}.

\begin{table*}
	\caption{Comparison results on Track 1 shows}
% 	\small
	\centering
	\begin{tabular}{p{2.2cm} c c c c c c c c c c c c c}
        \toprule
        	\multicolumn{14}{c}{\textbf{VALIDATION SET (Track 1 Shows)}}\\
            \midrule
        
        	\multirow{2}{*}{Method} & \multirow{2}{*}{Similarity kernel} & \multicolumn{6}{c}{hit@k} & \multicolumn{6}{c}{recall@k} \\
			
			\cmidrule(lr){3-8}  \cmidrule(lr){9-14}

			 & & k=5 & k=10 & k=20  & k=30 & k=40 & k=50 & k=50 & k=100 & k=200 & k=300 & k=400 & k=500 \\
		
			\midrule
%			\addlinespace[-0.6em]
			
			%&     \multicolumn{6}{c}{}     \\
			
%			[\defaultaddspace]
			\small{CBVRP baseline} \citep{cbvrp-acmmm-2018} & - & 0.253   & 0.347 & 0.442  & \textbf{0.510} & - & - & 0.111 & \textbf{0.175} & 0.264 & 0.329 & - & -\\

			\midrule
			\addlinespace[0.2em]
            
			OASIS \citep{OASIS} & Bilinear & 0.101  & 0.102 & 0.212 & 0.221  & 0.307 & 0.308 & 0.0215 & 0.026 & 0.065 & 0.098 & 0.108 & 0.127 \\
            
            \midrule
			\addlinespace[0.2em]
			
            \multirow{2}{2cm}{2-layer Neural Network} & Cosine & 0.088  & 0.130 & 0.192 & 0.244  & 0.282 & 0.307 & 0.041 & 0.062 & 0.094 & 0.122 & 0.144 & 0.164 \\
            & RBF & 0.067  & 0.088 & 0.144 & 0.187  & 0.234 & 0.261 & 0.027 & 0.044 & 0.076 & 0.104 & 0.132 & 0.159 \\
            & Softmax & 0.068  & 0.106 & 0.155 & 0.188  & 0.219 & 0.245 & 0.026 & 0.041 & 0.069 & 0.098 & 0.126 & 0.151 \\
            
            \midrule
			\addlinespace[0.2em]
			\multirow{2}{2cm}{OMKS with delta features \citep{OMKS}} & Cosine  & 0.219   & 0.321 & 0.431  & 0.452 & 0.481 & 0.521 & 0.109 & 0.162 & 0.217 & 0.268 & 0.301 & 0.319
            \\ & RBF & 0.230   & 0.318 & 0.402  & 0.448 & 0.478 & 0.501 & 0.121 & 0.181 & 0.254 & 0.300 & 0.338 & 0.368
            \\ & Softmax & 0.221   & 0.307 & 0.419  & 0.437 & 0.471 & 0.514 & 0.113 & 0.174 & 0.235 & 0.289 & 0.317 & 0.341 \\
            
            \midrule
			\addlinespace[0.2em]
			\multirow{2}{2cm}{FusedLSTM} & Cosine  & 0.208 & 0.263 & 0.362  & 0.421 & 0.467 & 0.498 & 0.090 & 0.168 & 0.211 & 0.254 & 0.288 & 0.313 \\
               & RBF & 0.251 & 0.311 & 0.451  & 0.487 & 0.528 & 0.552 & 0.131 & 0.197 & 0.261 & 0.308 & 0.351 & 0.372 \\
               & Softmax & \textbf{0.265} & 0.343 & 0.435  & 0.483 & 0.522 & 0.545 & \textbf{0.139} & \textbf{0.205} & \textbf{0.277} & 0.327 & 0.364 & 0.397 \\
        \bottomrule
	\end{tabular}
    \begin{tabular}{p{2.5cm} c c c c c c c c c c c c c}
        \toprule
        	\multicolumn{14}{c}{\textbf{TESTING SET (Track 1 Shows)}}\\
            \midrule
        	
        	\multirow{2}{*}{Method} & \multirow{2}{*}{Similarity kernel} & \multicolumn{6}{c}{hit@k} & \multicolumn{6}{c}{recall@k} \\
			
			\cmidrule(lr){3-8}  \cmidrule(lr){9-14}

			 & & k=5 & k=10 & k=20  & k=30 & k=40 & k=50 & k=50 & k=100 & k=200 & k=300 & k=400 & k=500 \\
		
			\midrule
%			\addlinespace[-0.6em]
			
			%&     \multicolumn{6}{c}{}     \\
			
%			[\defaultaddspace]
			CBVRP \citep{cbvrp-acmmm-2018} & - & 0.234 & 0.328 & 0.444  & \textbf{0.510} & - & - & 0.079 & \textbf{0.132} & 0.206 & 0.257 & - & -\\

			\midrule
			\addlinespace[0.2em]
			OMKS \newline (Submission 1) & RBF & 0.217  & 0.290 & 0.372  & 0.423 & 0.463 & 0.493 & 0.073 & 0.113 & 0.164 & 0.202 & 0.237 & 0.267\\
            \midrule
			\addlinespace[0.2em]
			
            FusedLSTM \newline (Submission 2) & Softmax & 0.223 & 0.288 & 0.368  & 0.420 & 0.456 & 0.484 & 0.075 & 0.113 & 0.162 & 0.199 & 0.231 & 0.261 \\
        \bottomrule
	\end{tabular}
\end{table*}

\begin{table*}
	\caption{Comparison results on Track 2 movies}
% 	\small
	\centering
	\begin{tabular}{p{2.2cm} c c c c c c c c c c c c c}
        \toprule
        	\multicolumn{14}{c}{\textbf{VALIDATION SET (Track 2 Movies)}}\\
            \midrule
        
        	\multirow{2}{*}{Method} & \multirow{2}{*}{Similarity kernel} & \multicolumn{6}{c}{hit@k} & \multicolumn{6}{c}{recall@k} \\
			
			\cmidrule(lr){3-8}  \cmidrule(lr){9-14}

			 & & k=5 & k=10 & k=20  & k=30 & k=40 & k=50 & k=50 & k=100 & k=200 & k=300 & k=400 & k=500 \\
		
			\midrule
			\small{CBVRP baseline} \citep{cbvrp-acmmm-2018} & - & 0.167 & 0.213 & 0.300  & \textbf{0.366} & - & - & 0.101 & \textbf{0.143} & 0.206 & 0.257 & - & -\\

			\midrule
			\addlinespace[0.2em]
            
			OASIS \citep{OASIS} & Bilinear & 0.077  & 0.079 & 0.091 & 0.098  & 0.212 & 0.276 & 0.0215 & 0.026 & 0.065 & 0.098 & 0.108 & 0.127 \\
            
            \midrule
			\addlinespace[0.2em]
			\multirow{2}{2cm}{OMKS with delta features \citep{OMKS}} & Cosine  & 0.142 & 0.167 & 0.238  & 0.313 & 0.351 & 0.409 & 0.079 & 0.132 & 0.167 & 0.241 & 0.298 & 0.318 \\
             & RBF & 0.162 & 0.187 & 0.267  & 0.343 & 0.392 & 0.421 & 0.087 & 0.154 & 0.189 & 0.257 & 0.310 & 0.332 \\
             & Softmax & 0.174 & 0.193 & 0.285  & 0.353 & 0.382 & 0.415 & 0.092 & 0.143 & 0.197 & 0.278 & 0.311 & 0.327 \\
            
            \midrule
			\addlinespace[0.2em]
			\multirow{2}{2cm}{FusedLSTM} & Cosine  & 0.178 & 0.188 & 0.321  & 0.367 & 0.403 & 0.428 & 0.104 & 0.151 & 0.213 & 0.261  & 0.315 & 0.335 \\
               & RBF & 0.151 & 0.187 & 0.317  & 0.353 & 0.398 & 0.413 & 0.101 & 0.161 & 0.198 & 0.308 & 0.310 & 0.331 \\
               & Softmax & 0.165 & 0.193 & 0.315  & 0.383 & 0.392 & 0.425 & 0.112 & 0.173 & 0.207 & 0.281 & 0.314 & 0.337 \\
        \bottomrule
	\end{tabular}
    \begin{tabular}{p{2.2cm} c c c c c c c c c c c c c}
        \toprule
        	\multicolumn{14}{c}{\textbf{TESTING SET (Track 2 Movies)}}\\
            \midrule
        	
        	\multirow{2}{*}{Method} & \multirow{2}{*}{Similarity kernel} & \multicolumn{6}{c}{hit@k} & \multicolumn{6}{c}{recall@k} \\
			
			\cmidrule(lr){3-8}  \cmidrule(lr){9-14}

			 & & k=5 & k=10 & k=20  & k=30 & k=40 & k=50 & k=50 & k=100 & k=200 & k=300 & k=400 & k=500 \\
		
			\midrule
			CBVRP \citep{cbvrp-acmmm-2018} & - & 0.167   & 0.227 & 0.303  & \textbf{0.356} & - & - & 0.073 & \textbf{0.106} & 0.152 & 0.189 & - & -\\

			\midrule
			\addlinespace[0.2em]
			OMKS \citep{OMKS} & RBF & 0.159 & 0.211 & 0.281 & 0.327 & 0.367 & 0.395 & 0.068 & 0.096 & 0.138 & 0.169 & 0.193 & 0.217\\
            \midrule
			\addlinespace[0.2em]
			
            FusedLSTM & Softmax & 0.145 & 0.190 & 0.259  & 0.296 & 0.333 & 0.366 & 0.063 & 0.089 & 0.125 & 0.153 & 0.176 & 196 \\
        \bottomrule
	\end{tabular}
\end{table*}

\subsection{Results}
Table 1 and 2 below give a complete comparison of all the attempted methods. For the triplet networks, the results are reported on the best choice of embedding size (256) and after stoppage at epoch \#4.

\section{Conclusion}
The main contribution of this paper was to introduce a FusedLSTM module in the Triplet Network to tend to the temporal characteristics in the video relevance prediction. The analysis as tabulated, shows that the best model was Fused LSTM based network which performed significantly better than the baseline and the OMKS algorithm also gives at par performance. The analysis also reinforces the idea that LSTM based networks are better for understanding the video content. Due to time constraints, the model could not be trained on the validation set (for evaluation on the testing set). Hence, the performance on the testing set might appear to be less than the baseline.

\section{Future Work}
Ensemble methods have proven to be useful in improving the performances of various recommendation systems (RS) \citep{ensemble1, ensemble2}. It was shown in \citep{ensemble0} that ensembling  simple RS models can perform better than a single complex model. Hence, different ways of ensembling the methods discussed in this paper can be explored to improve on the existing performance on the CBVRP task.

\bibliographystyle{ACM-Reference-Format}
\bibliography{sample-bibliography}

\end{document}